\documentclass{article}

\usepackage[final]{neurips_2021}

\usepackage[utf8]{inputenc} 
\usepackage[T1]{fontenc}    
\usepackage{hyperref}       
\usepackage{url}            
\usepackage{booktabs}       
\usepackage{amsfonts}       
\usepackage{nicefrac}       
\usepackage{microtype}      
\usepackage{xcolor}         
\usepackage[pdftex]{graphicx}  
\usepackage[english]{babel}
\usepackage{amsmath}
\usepackage{multirow}
\usepackage{textcase}

\newcommand{\argmin}{\operatornamewithlimits{argmin}}

\newcommand{\argmax}{\operatornamewithlimits{argmax}}

\def\encoder{q_{\phi_{\varepsilon}}(\boldsymbol {\varepsilon} | \mathbf{x})}
\def\encodertrained{q_{\phi_{\varepsilon}}^{*}(\boldsymbol{\varepsilon} | \mathbf{x})}

\def\classifier{p_{\theta_{\mathrm{y}}}(\mathbf{y} | \boldsymbol{\varepsilon})}

\title{Information-theoretic stochastic contrastive conditional GAN: InfoSCC-GAN}

\author{Vitaliy Kinakh$^{1}$\thanks{V. Kinakh and S. Voloshynovskiy are corresponding authors.}, Mariia Drozdova$^{1,2}$, Guillaume Quétant$^{1,2}$, \\ \textbf{Tobias Golling$^{2}$ \& Slava Voloshynovskiy$^{1*}$} \\
  $^1$Department of Computer Science\\
  $^2$Department of Particle Physics\\
  University of Geneva\\
  Switzerland\\
  \texttt{\{vitaliy.kinakh,svolos\}@unige.ch} \\
}

\begin{document}

\maketitle

\begin{abstract}
Conditional generation is a subclass of generative problems where the output of the generation is conditioned by the attribute information. In this paper, we present a stochastic contrastive conditional generative adversarial network (InfoSCC-GAN) with an explorable latent space. The InfoSCC-GAN architecture is based on an unsupervised contrastive encoder built on the InfoNCE paradigm, an attribute classifier and an EigenGAN generator. We propose a novel training method, based on generator regularization using external or internal attributes every $n$-th iteration, using a pre-trained contrastive encoder and a pre-trained classifier. The proposed InfoSCC-GAN is derived based on an information-theoretic formulation of mutual information maximization between input data and latent space representation as well as latent space and generated data. Thus, we demonstrate a link between the training objective functions and the above information-theoretic formulation. The experimental results show that InfoSCC-GAN outperforms the "vanilla" EigenGAN in the image generation on AFHQ and CelebA datasets. In addition, we investigate the impact of discriminator architectures and loss functions by performing ablation studies.
Finally, we demonstrate that thanks to the EigenGAN generator, the proposed framework enjoys a stochastic generation in contrast to vanilla deterministic GANs yet with the independent training of encoder, classifier, and generator in contrast to existing
frameworks.
Code, experimental results, and demos are available online at \href{https://github.com/vkinakh/InfoSCC-GAN}{github.com/vkinakh/InfoSCC-GAN}. %

\end{abstract}

\section {Introduction}
\label{introduction}

In this paper, we present a new information-theoretic stochastic contrastive conditional generative adversarial network InfoSCC-GAN. The proposed approach is based on the stochastic generative model EigenGAN \cite{He2021EigenGANLE} with explorable latent spaces, independent contrastive encoder and independent classifier for class label regularization. The EigenGAN baseline generator ensures that the model is truly stochastic. In contrast to other conditional generation methods, our model is based on an independent contrastive encoder and attribute classifier. By using them, we avoid a complex and joint procedure of encoder and classifier training, when the model does not produce realistic images in the early iterations. Also, since we use the encoder pre-trained on the real data, we ensure that it properly contrasts real data and avoids contrasting poorly generated data. 

We provide an information-theoretical problem formulation of the proposed model in Section \ref{formulation}.


We summarize our contributions in this paper as follows: (i) we proposed a novel stochastic contrastive conditional generative adversarial network (InfoSCC-GAN) for stochastic conditional image generation with explorable latent space. It is based on the EigenGAN generator, an independent contrastive encoder and an independent attributes' classifier; (ii) we introduce a novel classification regularization technique, which is based on updating the model each $n$-th iteration and updating the generator using adversarial and classification loss separately; (iii) we provide the information-theoretic interpretation of the proposed system; (iv) we perform the ablation studies to determine the contribution of each part of the model to the overall performance.

    
    
    

\section {Information-theoretic formulation}
\label{formulation}

\begin{figure}
\centering
    \includegraphics[width=0.7\linewidth]{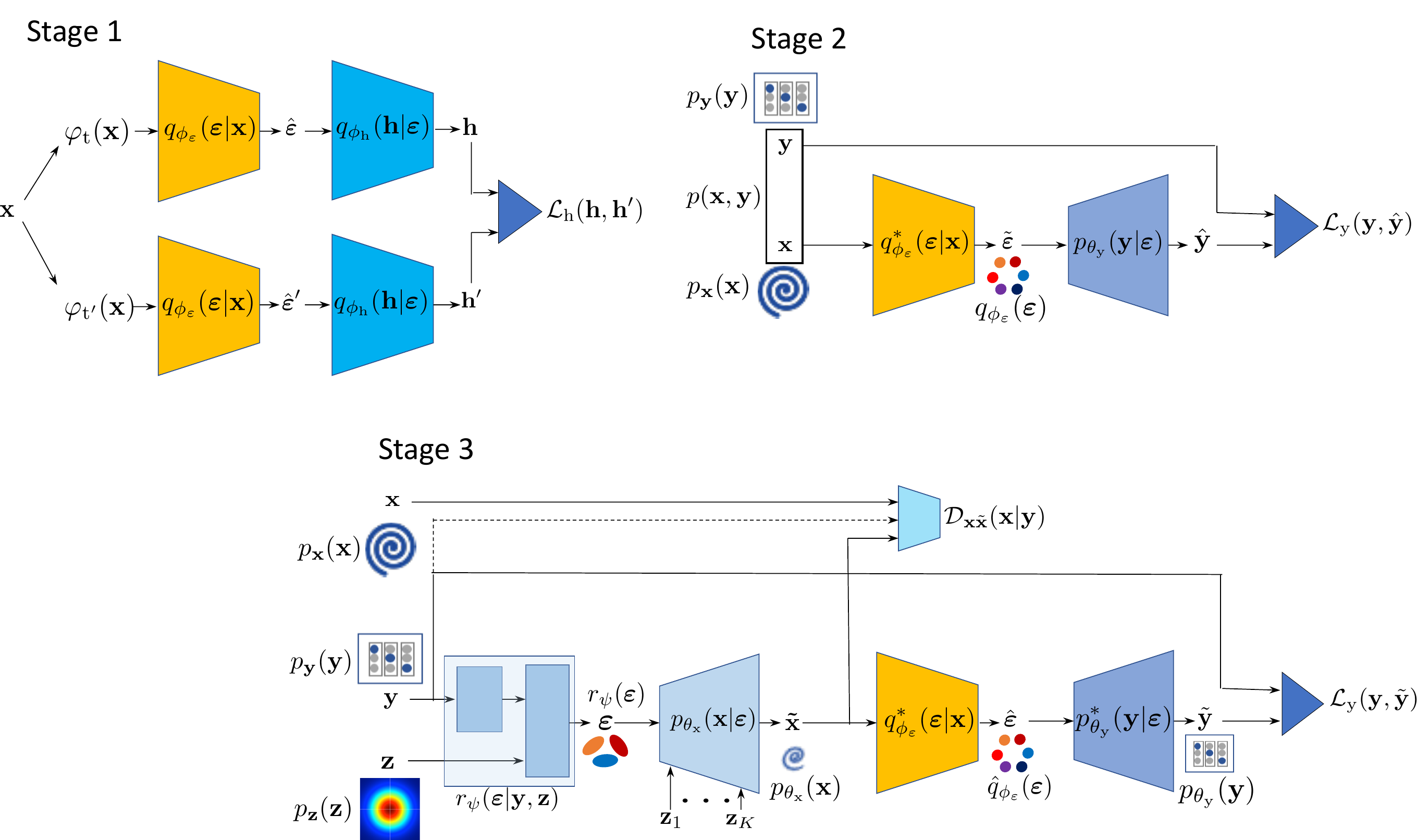}
    \caption{The proposed InfoSCC-GAN. Stage 1. Contrastive encoder training. Stage 2. Classifier training. Stage 3. Conditional generator training.}
\label{proposed_scheme}
\end{figure}

The training stages of InfoSCC-GAN are shown in Fig. \ref{proposed_scheme} and explained below.

\subsection{The training of the encoder (stage 1)}

The encoder training is based on the maximization problem:

\begin{equation}
  \label{MI_SIMCLR}
    \hat{\boldsymbol{\phi_{\varepsilon}}} = \argmax_{\boldsymbol{\phi_{\varepsilon}}} I_{\boldsymbol {\phi_{\varepsilon}}}({\bf{X}} ;{\bf{E}}), 
\end{equation}

where $I_{{\boldsymbol \phi}_{\varepsilon}}({\bf X} ;{\bf E}) =  \mathbb{E}_{p(\mathbf{x} , {\boldsymbol\varepsilon}) }\left[   \log \frac{q_{ {\boldsymbol \phi}_{\varepsilon}}({\boldsymbol \varepsilon} | {\bf x} )}{q_{ {\boldsymbol \phi}_{\varepsilon}}({\boldsymbol \varepsilon})}   \right]$, where $\encoder$ denotes the encoder and $q_{\phi_{\varepsilon}}(\boldsymbol \varepsilon)$ - the marginal latent space distribution.

In the framework of contrastive learning, (\ref{MI_SIMCLR}) is maximized based on the infoNCE framework \cite{Oord2018RepresentationLW}. In the practical implementation, one can use the approach similar to SimCLR\cite{Chen2020ASF}, where the inner product between the positive pairs created from the augmented views of the same image is maximized and the inner product between the negative pairs based on different images is minimized\footnote{The SimCLR training is based on the maximization $I_{\boldsymbol \phi_{\varepsilon}, \phi_{n}}({\bf X} ;{\bf H})$, but since $I_{\boldsymbol \phi_{\varepsilon}, \phi_{n}}({\bf X} ;{\bf H}) < I_{\boldsymbol \phi_{\varepsilon}}({\bf X} ;{\bf E})$ one could lower bound (\ref{MI_SIMCLR}).}. Alternatively, one can use other approaches to learn the representation $\boldsymbol \varepsilon$ such as BYOL \cite{Grill2020BootstrapYO}, Barlow Twins \cite{Zbontar2021BarlowTS}, etc. without loose of generality of the proposed approach. It should be pointed out that the encoder is trained independently from the decoder in the scope of the considered setup.

\subsection{ The training of the class attribute classifier (stage 2)}

The class attribute classifier training is based on the maximization problem: 

\begin{equation}
\label{MI_Classifier}
\hat{\boldsymbol \theta}_{\mathrm{y}} = \argmax_{ \boldsymbol \theta_{\mathrm y}} I_{{\boldsymbol \phi}^*_{\varepsilon},{\boldsymbol \theta}_{\mathrm y} }({\bf Y} ;{\bf E}), 
  \end{equation}
where $I_{{\boldsymbol \phi}^*_{\varepsilon},{\boldsymbol \theta}_{\mathrm y} }({\bf Y} ;{\bf E}) = H({\bf Y}) - H_{{\boldsymbol \phi}^*_{\varepsilon},{\boldsymbol \theta}_{\mathrm y} }({\bf Y}|{\bf E})$ and $H({\bf Y})  = - \mathbb{E}_{p_{\bf y}({\bf y})}\log p_{\bf y}({\bf y}) $ is the conditional entropy of ${\bf Y}$ and the conditional entropy is defined as 
 $ H_{{\boldsymbol \phi}^*_{\varepsilon},{\boldsymbol \theta}_{\mathrm y} }({\bf Y}|{\bf E}) = - \mathbb{E}_{p_{\bf x}({\bf x})}  \left[   \mathbb{E}_{q_{{\boldsymbol \phi}^*_{\varepsilon}}({\boldsymbol \varepsilon}| \mathbf{x})} \left[  \log p_{{\boldsymbol \theta}_{\mathrm{y}}}(\mathbf{y} |{ \boldsymbol \varepsilon}) \right]\right]$. Since $H({\bf Y})$ is independent of the parameters of the encoder and classifier, (\ref{MI_Classifier}) reduces to the lower bound minimization:
\begin{equation}
\label{MI_Classifier_min}
\hat{\boldsymbol \theta}_{\mathrm y} = \argmin_{ {\boldsymbol \theta}_{\mathrm y}} H_{{\boldsymbol \phi}^*_{\varepsilon},{\boldsymbol \theta}_{\mathrm y} }({\bf Y}|{\bf E}) , 
  \end{equation}
that under the categorical conditional distribution $p_{{\boldsymbol \theta}_{\mathrm{y}}}(\mathbf{y} |{ \boldsymbol \varepsilon})$ can be expressed as the categorical cross entropy $\mathcal{L}_{\mathrm{y}}(\mathbf{y}, \hat{\mathbf{y}})$. 

\subsection{ The training of the decoder, i.e., the mapper and generator (stage 3) }

The decoder is trained first to maximize the mutual information between the class attributes $\tilde{ \bf y}$ predicted from the generated images and true class attributes $\bf y$:

\begin{equation}
\label{MI_decoder_classification_loss}
(\hat{\boldsymbol \theta}_{\mathrm x},\hat{\boldsymbol \psi} ) = \argmax_{ \boldsymbol \theta_{\mathrm x}, {\boldsymbol \psi}} I_{{\boldsymbol \psi},{\boldsymbol \theta}_{\mathrm x},{\boldsymbol \phi}^*_{\varepsilon},{\boldsymbol \theta}^*_{\mathrm y} }({\bf Y} ;{\bf E}), 
  \end{equation}
where $I_{{\boldsymbol \psi},{\boldsymbol \theta}_{\mathrm x},{\boldsymbol \phi}^*_{\varepsilon},{\boldsymbol \theta}^*_{\mathrm y} }({\bf Y} ;{\bf E}) = H({\bf Y}) - H_{{\boldsymbol \psi},{\boldsymbol \theta}_{\mathrm x},{\boldsymbol \phi}^*_{\varepsilon},{\boldsymbol \theta}^*_{\mathrm y} }({\bf Y}|{\bf E})$ and $H({\bf Y})  = - \mathbb{E}_{p_{\bf y}({\bf y})}\log p_{\bf y}({\bf y}) $ and the conditional entropy is defined as 
 $ H_{{\boldsymbol \psi},{\boldsymbol \theta}_{\mathrm x},{\boldsymbol \phi}^*_{\varepsilon},{\boldsymbol \theta}^*_{\mathrm y} }({\bf Y}|{\bf E}) = - \mathbb{E}_{p_{\bf y}({\bf y})}  \left[ \mathbb{E}_{p_{\bf z}({\bf z})}  \left[   \mathbb{E}_{r_{{\boldsymbol \psi}}({\boldsymbol \varepsilon}| \mathbf{y}, {\bf z})}  \left[     \mathbb{E}_{p_{{\boldsymbol \theta}_{\mathrm x}}(\mathbf{x}|{\boldsymbol \varepsilon})}   
 \left[  \mathbb{E}_{q_{{\boldsymbol \phi}^*_{\varepsilon}}({\boldsymbol \varepsilon}| \mathbf{x})}        
       \left[     \log p_{{\boldsymbol \theta}^*_{\mathrm{y}}}(\mathbf{y} |{ \boldsymbol \varepsilon}) \right]\right]\right] \right]\right]$, $\classifier$ corresponds to the classifier and $\encodertrained$ denotes the pre-trained encoder. Since $H({\bf Y})$ is independent of the parameters of the encoder and classifier, (\ref{MI_decoder_classification_loss}) reduces to the lower bound minimization:
\begin{equation}
\label{MI_decoder_classification_loss_bound}
(\hat{\boldsymbol \theta}_{\mathrm x},\hat{\boldsymbol \psi} ) = \argmin_{ \boldsymbol \theta_{\mathrm x}, {\boldsymbol \psi}}  H_{{\boldsymbol \psi},{\boldsymbol \theta}_{\mathrm x},{\boldsymbol \phi}^*_{\varepsilon},{\boldsymbol \theta}^*_{\mathrm y} }({\bf Y}|{\bf E}),
  \end{equation}
that under the categorical conditional distribution $p_{{\boldsymbol \theta}_{y}}(\mathbf{y} |{ \boldsymbol \varepsilon})$ can be expressed as the categorical cross entropy $\mathcal{L}_{\mathrm{y}}(\mathbf{y}, \tilde{\mathbf{y}})$. 

Finally, the decoder should produce samples that follow the distribution of training data $p_{\bf x}({\bf x})$ that corresponds to the maximization of mutual information:  

\begin{equation}
\label{MI_decoder_generation_loss}
(\hat{\boldsymbol \theta}_{\mathrm x},\hat{\boldsymbol \psi} ) = \argmax_{ \boldsymbol \theta_{\mathrm x}, {\boldsymbol \psi}} I_{{\boldsymbol \psi},{\boldsymbol \theta}_{\mathrm x} }({\bf X} ;{\bf E}), 
  \end{equation}
where $ I_{{\boldsymbol \psi},{\boldsymbol \theta}_{\mathrm x} }({\bf X} ;{\bf E}) = \mathbb{E}_{p_{\bf x}({\bf x})}  \left[       \mathbb{E}_{p_{\bf y}({\bf y})}  \left[ \mathbb{E}_{p_{\bf z}({\bf z})}  \left[   \mathbb{E}_{r_{{\boldsymbol \psi}}({\boldsymbol \varepsilon}| \mathbf{y}, {\bf z})}  \left[     \mathbb{E}_{p_{{\boldsymbol \theta}_{\mathrm x}}(\mathbf{x}|{\boldsymbol \varepsilon})}       
       \left[      \log \frac{p_{ {\boldsymbol \theta}_{\mathrm x}}( {\bf x}|{\boldsymbol \varepsilon}  )}{p_{ {\bf x}}( {\bf x}  )}   \right]\right]\right] \right]   \right]        =   \mathbb{E}_{p_{\bf y}({\bf y})}  \left[ \mathbb{E}_{p_{\bf z}({\bf z})}  \left[   \mathbb{E}_{r_{{\boldsymbol \psi}}({\boldsymbol \varepsilon}| \mathbf{y}, {\bf z})}  
       \left[    \mathbb{D}_{\mathrm{KL}}(p_{ {\boldsymbol \theta}_{\mathrm x}}( {\bf x}|{{\bf E} =\boldsymbol \varepsilon}  )||p_{ {\boldsymbol \theta}_{\mathrm x}}( {\bf x} ))   \right]\right] \right]    -    \mathbb{D}_{\mathrm{KL}}(p_{\bf x}({\bf x})||p_{ {\boldsymbol \theta}_{\mathrm x}}( {\bf x} ))  $, where $p_{\theta_{\mathrm{x}}}(\bf x)$ denotes the distribution of generated samples $\bf{\tilde{x}}$. Since $ \mathbb{D}_{\mathrm{KL}}(p_{ {\boldsymbol \theta}_{\mathrm x}}( {\bf x}|{{\bf E} =\boldsymbol \varepsilon}  )||p_{ {\boldsymbol \theta}_{\mathrm x}}( {\bf x} ))   \geq 0$, the maximization of the above mutual information reduces to the minimization:

\begin{equation}
\label{MI_decoder_discriminnator}
(\hat{\boldsymbol \theta}_{\mathrm x},\hat{\boldsymbol \psi} ) = \argmin_{ \boldsymbol \theta_{\mathrm x}, {\boldsymbol \psi}} \mathbb{D}_{\mathrm{KL}}(p_{\bf x}({\bf x})||p_{ {\boldsymbol \theta}_{\mathrm x}}( {\bf x} )). 
  \end{equation}
The above discriminator is denoted as $\mathcal{D}_{\mathbf{x} \tilde{\mathbf{x}}}(\mathbf{x})$. At the same time, one can also envision the discriminator conditioned on the attribute class $\bf y$, e.g., $\mathcal{D}_{\mathbf{x} \tilde{\mathbf{x}}}(\mathbf{x} \mid \mathbf{y})$ that is implemented as a set of discriminators for each subset of generated and original samples defined by the class attributes $\bf y$.

\section{Experiments}


In this section, we describe the generation experiments. For the evaluation, we use 3 metrics: Fréchet inception distance (FID) \cite{Heusel2017GANsTB}, inception score (IS) \cite{Salimans2016ImprovedTF} and Chamfer distance \cite{ravi2020pytorch3d}. Since Chamfer distance works in low dimensional spaces, we compute features of the real and generated image by the pre-trained encoder, then compute the 3D t-SNEs of these features, which are used to compute the Chamfer distance. We perform ablation studies on AFHQ dataset. To determine whether the conditional generated images obey the needed attributes, we use attribute control accuracy. The attribute control accuracy is computed as the percentage of the images for which the output of the attribute classifier is the same as an input attribute. The attribute control accuracy measures how good the generator is at conditional generation.

\subsection{EigenGAN}

We compare the proposed InfoSCC-GAN with the original EigenGAN \cite{He2021EigenGANLE} on the AFHQ dataset. Our model is based on the same generator while using different inputs and conditional regularization. In the current setup, EigenGAN has 6 layers each with 6 dimensions that are used for interpretable and controllable features exploration. The original EigenGAN achieves FID score of \textbf{29.02} and IS of \textbf{8.52} after 200000 training iterations on AFHQ dataset using global discriminator and Hinge loss \cite{Lim2017GeometricG}. The EigenGAN does not allow for interpretable feature exploration for the wild animal images. It can be explained by the imbalance since the "wild" animals class includes multiple distinct subclasses such as tiger, lion, fox, and wolf, which are not semantically close.

\subsection{Conditional generation}

We achieve the best FID score of \textbf{11.59}, IS of \textbf{11.06} and Chamfer distance \textbf{3645} using the InfoSCC-GAN approach after 200000 training iterations using Patch discriminator \cite{Isola2017ImagetoImageTW} and LSGAN \cite{Mao2017LeastSG} loss on AFHQ dataset. In the current setup, we have 6 layers with 6 explorable dimensions. 
The results on CelebA dataset with 5, 10 and 15 attribute labels are presented in Tables \ref{table:table_resuls_CelebA_5}, \ref{table:table_results_CelebA_10}, \ref{table:table_results_CelebA_15}.
\begin{table}[]
\centering
\caption{Conditional generation results on CelebA dataset with 5 selected attributes.}
\label{table:table_resuls_CelebA_5}
\begin{tabular}{ccccccc}
\hline
\multirow{2}{*}{\textbf{FID $\downarrow$}} & \multirow{2}{*}{\textbf{IS $\uparrow$}}  & \multicolumn{5}{c}{\textbf{Attribute Control Accuracy $\uparrow$}} \\ 
\cline{3-7}  &  & Bald & Eyeglasses & Mustache & Wearing Hat & Wearing Necktie \\ \hline
\multicolumn{1}{r}{27.84} & \multicolumn{1}{r}{9.91} & \multicolumn{1}{r}{93.27\%} & \multicolumn{1}{r}{99.88\%} & \multicolumn{1}{r}{95.68\%} & \multicolumn{1}{r}{94.62\%} & \multicolumn{1}{r}{98.62\%} \\ \hline
\end{tabular}
\end{table}

\begin{table}[]
\caption{Conditional generation results on CelebA dataset with 10 selected attributes.}
\label{table:table_results_CelebA_10}
\begin{tabular}{ccccccc}
\hline
\multirow{2}{*}{\textbf{FID$\downarrow$}} & \multirow{2}{*}{\textbf{IS$\uparrow$}} & \multicolumn{5}{c}{\textbf{Attribute Control Accuracy$\uparrow$}}  \\ \cline{3-7} 
 &  & Bald & Black Hair & Blond Hair & Brown Hair & Double Chin \\ \hline
32.39 & 9.04 & 89.74\% & 89.61\% & 86.86\% & 85.55\% & 84.82\% \\ \hline
\multicolumn{1}{l}{} & \multicolumn{1}{l}{} & Eyeglasses & Gray Hair & Mustache & Wearing Hat & Wearing Necktie \\ \cline{3-7} 
\multicolumn{1}{l}{} & \multicolumn{1}{l}{} & 99.6\% & 81.71\% & 92.27\% & 92.83\% & 89.26\% \\ \cline{3-7} 
\end{tabular}
\end{table}

\begin{table}[]
\caption{Conditional generation results on CelebA dataset with 15 selected attributes.}
\label{table:table_results_CelebA_15}
\begin{tabular}{ccccccc}
\hline
\multirow{2}{*}{\textbf{FID$\downarrow$}} & \multirow{2}{*}{\textbf{IS$\uparrow$}} & \multicolumn{5}{c}{\textbf{Attribute Control Accuracy$\uparrow$}}\\ \cline{3-7} 
& & Bald & Blurry & Chubby & Double Chin & Eyeglasses \\ \hline
34.97 & 8.87 & 83.6\%  & 96.46\% & 80.1\% & 95.74\% & 98.11\% \\ \hline
& & Goatee & Gray Hair & Mustache & Narrow Eyes & Pale Skin \\ \cline{3-7} 
& & 89.09\% & 90.78\% & 87.64\% & 74.22\% & 86.91\% \\ \cline{3-7} 
& & Receding Hairline & Rosy Chicks & Sideburns & Wearing Hat & Wearing Necktie \\ \cline{3-7} 
&  & 86.46\% & 78.88\% & 74.9\% & 97.64\% & 94.87\% \\ \cline{3-7} 
\end{tabular}
\end{table}

\section{Ablation studies}


In this section, we describe the ablation studies we have performed on the type of discriminator and the discriminator loss.

\subsection{Discriminator ablation studies}

\begin{table}
\centering
\caption{Discriminator ablation studies.}
\label{table:table_discriminator_ablation}
\begin{tabular}{llrrr}
\hline
\textbf{Discriminator} & \textbf{Loss}  & \textbf{FID $\downarrow$} & \textbf{IS $\uparrow$} & \textbf{Chamfer distance $\downarrow$} \\ \hline
Global & Hinge          & 13.08          & 10.71          & 4030           \\ \hline
Global & Non saturating & 25.62          & 10.33          & 28595          \\ \hline
Global & LSGAN          & 29.02          & 9.89           & 45583          \\ \hline
Patch  & Hinge          & 15.95          & 10.51          & 7327           \\ \hline
Patch  & Non saturating & 14.83          & 10.21          & 5114           \\ \hline
Patch  & LSGAN          & \textbf{11.59} & \textbf{11.06} & \textbf{3645}  \\ \hline
\end{tabular}
\end{table}

In this section, we describe the discriminator and loss ablation studies. We compare two discriminators: global discriminator and patch discriminator. The global discriminator outputs one value that is the probability of the image being real. The architecture of the global discriminator is inspired by the EigenGAN paper. The patch discriminator outputs a tensor of values that represent the probability of the image patch being real, the architecture of the patch discriminator is inspired by the pix2pix GAN \cite{Isola2017ImagetoImageTW}. We compare these discriminators in combination with discriminator losses: Hinge loss, non-saturating loss and LSGAN. The results of the studies are presented in Table. \ref{table:table_discriminator_ablation}.  For all of the discriminators and losses, used in the study, the attribute control accuracy is in the range of 99-100\%. 

\section{Conclusions}

In this paper, we propose a novel stochastic contrastive conditional GAN InfoSCC-GAN, which produces stochastic conditional image generation with an explorable latent space. We provide the information-theoretical formulation of the proposed system. Unlike other contrastive image generation approaches, our method is truly a stochastic generator, that is controlled by the class attributes and by the set of stochastic parameters. We apply a novel training methodology based on using a pre-trained unsupervised contrastive encoder and a pre-trained classifier with every $n$-th iteration using a classification regularization. We propose an information-theoretical interpretation of the proposed system. We propose a novel attribute selection approach based on clustering embeddings computed using an encoder. The proposed model outperforms "vanilla" EigenGAN on AFHQ dataset, while it also provides conditional image generation. We have performed ablations studies to determine the best setup for conditional image generation. Finally, we have performed experiments on AFHQ and CelebA datasets.


\begin{ack}
This research was partially funded by the SNF Sinergia project (CRSII5-193716) Robust deep density models for high-energy particle physics and solar flare analysis (RODEM).
\end{ack}

{
\small

\bibliographystyle{unsrtnat}
\bibliography{egbib}
}




\end{document}